\documentclass[11pt]{article}

\usepackage[utf8]{inputenc}
\usepackage[T1]{fontenc}
\usepackage[a4paper,margin=1in]{geometry}
\usepackage{booktabs}
\usepackage{array}
\usepackage{enumitem}
\usepackage{graphicx}
\usepackage{float}
\usepackage{microtype}
\usepackage{xcolor}
\definecolor{linkcolor}{HTML}{1F4E79}
\PassOptionsToPackage{hyphens}{url}
\usepackage[
  colorlinks=true,
  linkcolor=linkcolor,
  citecolor=linkcolor,
  urlcolor=linkcolor,
  breaklinks=true,
  pdftitle={A Reproducible Universal Dependencies-Style Pipeline for Katharevousa Greek Parliamentary Text},
  pdfauthor={George Mikros and Fotios Fitsilis}
]{hyperref}
\newcommand{\repobase}{https://github.com/gmikros/katharevousa-nlp-tooling}
\newcommand{\repofile}[1]{\href{\repobase/blob/main/#1}{\nolinkurl{#1}}}
\newcommand{\repodir}[1]{\href{\repobase/tree/main/#1}{\nolinkurl{#1}}}

\title{A Reproducible Universal Dependencies-Style Pipeline for
Katharevousa Greek Parliamentary Text\thanks{%
Preprint: \href{https://arxiv.org/abs/2605.22978}{arXiv:2605.22978}.
Source code, reference annotations, fixed splits, and benchmark reports are
openly released at \url{https://github.com/gmikros/katharevousa-nlp-tooling}.
A model release for the XLM-R configuration is planned on the Hugging Face
Hub under the project name \texttt{kathnlp}.}}

\author{
George Mikros\\
Hamad Bin Khalifa University
\and
Fotios Fitsilis\\
Universidad Austral
}

\date{}

\begin{document}
\maketitle

\begin{abstract}
Katharevousa Greek remains poorly served by contemporary NLP pipelines despite its importance for legal, administrative, and parliamentary archives. We present a reproducible workflow for building and evaluating a Universal Dependencies-style parsing resource for Katharevousa parliamentary questions from Greece's early post-junta period. The pipeline links OCR-aware reconstruction, schema-constrained LLM-assisted annotation, automatic validation, deterministic CoNLL-U snapshotting, fixed-split evaluation, and model-family comparison. The frozen automatically validated reference set contains 1{,}697 sentences, split into 1{,}357 training sentences and 340 held-out test sentences. We compare off-the-shelf Greek and Ancient Greek parsers, a feature-based parser, mBERT, XLM-R, and custom Stanza training under the same scoring protocol. Off-the-shelf systems show substantial register mismatch: the strongest external baseline, spaCy Greek, reaches 0.4183 LAS. The best structural parser, an XLM-R model, reaches 0.8893 UPOS accuracy, 0.7250 dependency-relation F1, 0.6098 UAS, and 0.5162 LAS, an absolute LAS gain of 0.0980 over the best external baseline. The feature-based model remains competitive for UPOS and relation labeling, indicating that transparent lexical-context features still matter at this data scale. Beyond scores, the paper contributes an auditable methodology for turning difficult historical parliamentary OCR into reusable syntactic NLP infrastructure. The entire pipeline---code, schema, frozen reference annotations, fixed train/test split, and per-model benchmark reports---is released as an open-access companion to this paper.
\end{abstract}


\section{Introduction}

Large multilingual NLP systems have expanded syntactic analysis to many languages, but historical registers often remain outside their effective coverage. Katharevousa Greek is a particularly revealing case. It is not Ancient Greek, but neither is it contemporary Demotic Greek: it combines official modern institutional vocabulary with archaizing morphology, legal formulae, and syntactic constructions that are weakly represented in current Greek NLP training data. This mismatch matters for digital archives. Greek parliamentary records from the democratic transition after the military junta contain written questions that are historically important, linguistically formal, and computationally difficult. Prior digital-humanities work on this archive showed that computational analysis can recover party style, topic structure, and register variation from OCR-derived written parliamentary questions \cite{mikrosfitsilis2025dsh}. The next bottleneck is syntactic infrastructure: without a reliable parser, downstream work on argument structure, institutional actors, policy claims, and diachronic register change remains fragile.

This paper develops a reproducible dependency parsing pipeline for Katharevousa parliamentary text and releases it as an open-access companion library, \texttt{kathnlp}, at \url{\repobase}. The contribution is not only a model score. It is a governed research workflow that converts noisy OCR-derived material into a fixed, inspectable reference snapshot and evaluates multiple model families under a shared protocol. We make four contributions:

\begin{enumerate}[leftmargin=1.5em]
  \item We define an end-to-end reconstruction and automatically validated annotation workflow for Katharevousa Greek parliamentary text, using UD-style CoNLL-U output with sidecar metadata for Katharevousa-specific lexical, orthographic, and register phenomena.
  \item We release a fixed evaluation protocol for a 1{,}697-sentence reference set, including deterministic train/test splits and benchmark reports for external, feature-based, transformer, and Stanza-based models.
  \item We show empirically that existing Greek and Ancient Greek parsers do not transfer cleanly to Katharevousa parliamentary syntax, while a small custom XLM-R parser improves LAS by 0.0980 over the best external baseline.
  \item We package the pipeline as an open-source Python project (\texttt{kathnlp}) so that every empirical claim in this paper is traceable to a script, a configuration file, and a benchmark report inside the released repository.
\end{enumerate}

The results are intentionally reported with both positive and negative optimization outcomes. This is important in low-resource historical NLP: the path to a usable parser includes not only the winning configuration, but also the failed schedules, framework constraints, and validation decisions that make the final result reproducible.

\section{Background and Related Work}

\subsection{Katharevousa and parliamentary archives}

Katharevousa was an archaizing official register of Greek that mediated between modern public administration and classical linguistic prestige. In parliamentary and legal settings it functioned as a marker of institutional authority, but it also indexed the political language question that divided conservative, liberal, and left-wing traditions in modern Greece \cite{mackridge2009}. The written parliamentary questions used here come from the early Third Hellenic Republic, after the restoration of democracy in 1974 and around the 1976 language reform that replaced Katharevousa as the official state language. The previous companion study describes the broader archive: 1{,}338 OCR-derived written questions from 1976--1977, filtered to 1{,}009 validated texts and a 646-document analytic sample for party-level stylometry and topic modeling \cite{mikrosfitsilis2025dsh}. That work motivates the present NLP task: syntactic tools are needed to move beyond surface features toward reusable analysis of actors, relations, and institutional discourse.

\subsection{Universal Dependencies and low-resource parsing}

Universal Dependencies (UD) provides a cross-linguistically consistent representation for tokenization, morphology, and dependency syntax \cite{nivre2020ud}. CoNLL-U gives a stable exchange format for sentence-level comments and token-level annotation fields such as FORM, LEMMA, UPOS, FEATS, HEAD, and DEPREL \cite{udformat}. UD is attractive for Katharevousa because it lets us compare modern Greek, Ancient Greek, and custom models under a shared representation. The same choice also exposes a hard transfer problem: Katharevousa shares traits with both modern and pre-modern Greek, but the parliamentary register is not identical to either resource family.

Low-resource dependency parsing has long relied on cross-lingual transfer, multilingual representations, and careful exploitation of small treebanks \cite{duong2015lowresource,wu2018multilingual,kondratyuk2019udify}. Multilingual encoders such as mBERT \cite{devlin2019bert} and XLM-R \cite{conneau2020xlmr} are natural candidates because they encode Greek with broader cross-lingual supervision. However, transfer alone does not guarantee success when the target register contains OCR noise, archaizing morphology, and institutional formulae. We therefore compare pretrained systems, custom neural models, and a transparent feature-based baseline rather than assuming that a transformer will dominate all metrics.

\subsection{Historical OCR and constrained annotation}

Historical NLP pipelines often fail before modeling begins: OCR errors, line-break artifacts, historical spelling, and inconsistent metadata can distort the unit of analysis \cite{richter2018ocr,makarov2020normalization}. The companion DSH manuscript reports the archival funnel and the broader computational-stylistic workflow, including polytonic normalization, party metadata enrichment, and human-validated LLM assistance for metadata and topic labels \cite{mikrosfitsilis2025dsh}; the underlying corpus was produced by a dedicated machine-learning OCR digitisation of the parliamentary archive \cite{fitsilis2024digitization}. The present work uses LLMs differently: as constrained annotation infrastructure for UD-style syntactic records. Because LLM outputs can drift from schema, produce malformed JSON, or silently violate tree constraints, we treat them as annotation aids embedded inside validation, retry, and deterministic freezing steps. The resulting automatically validated snapshot serves as the fixed reference set for the experiments in this paper.

\section{Data}

\subsection{Source archive}

The source material consists of written parliamentary questions from 1976--1977, a period in which Greek parliamentary oversight was being rebuilt after authoritarian rule. The documents were digitised from the historical archive of the 1st Parliamentary Term (1974--1977) of the Hellenic Parliament: 1{,}674 page images corresponding to 1{,}338 questions were processed with a custom OCR platform combining YOLOv5-based text-line segmentation and a Calamari-OCR recognition engine trained on validated polytonic Greek, reaching a character recognition accuracy of 98.7\% (character error rate 1.27\%) \cite{fitsilis2024digitization}. The documents are therefore OCR-derived and belong to the same archive studied in the companion digital-humanities work \cite{mikrosfitsilis2025dsh}. That archive combines document degradation, OCR uncertainty, party metadata gaps, polytonic and monotonic orthographic variation, and a register that falls between Ancient and Modern Greek NLP resources. For syntactic modeling, the released pipeline reconstructs answer-file text from exported document and spreadsheet sources, then prepares a sentence-level annotation set.

\subsection{Reconstruction}

The repository encodes the reconstruction workflow as executable scripts (\repodir{scripts}):

\begin{itemize}[leftmargin=1.5em]
  \item source export from DOCX and XLSX inputs (\repofile{scripts/export_sources.py});
  \item OCR-aware reconstruction of answer-file text from spreadsheet columns (\repofile{scripts/reconstruct_answer_files.py});
  \item preprocessing of line-break hyphenation, split-word joins, punctuation around sentence boundaries, and long enumerative sentences;
  \item deterministic freezing of validated annotation batches into a final CoNLL-U snapshot (\repofile{scripts/freeze_final_gold_snapshot.py}).
\end{itemize}

This design separates archival reconstruction from model training. The final treebank is not a transient output of one experiment; it is a versioned artifact with a manifest, allowing later models to reuse the same split and scoring protocol.

\subsection{Reference snapshot}

The frozen reference set contains 1{,}697 sentences. Of these, 1{,}565 come from the main batch path and 132 from retry replacements. The fixed split uses seed 42 and contains 1{,}357 training sentences and 340 held-out test sentences. The held-out set contains 4{,}093 tokens. Table~\ref{tab:data} summarizes the data used for modeling. The snapshot and its manifest are released as \repofile{data/processed/final_gold/gold_final.conllu} and \repofile{data/processed/final_gold/snapshot_manifest.json}.

\begin{table}[H]
\centering
\caption{Reference snapshot and fixed split.}
\label{tab:data}
\begin{tabular}{lr}
\toprule
Item & Count \\
\midrule
Frozen reference sentences & 1{,}697 \\
Batch-origin sentences & 1{,}565 \\
Retry-replaced sentences & 132 \\
Training sentences & 1{,}357 \\
Held-out test sentences & 340 \\
Held-out test tokens & 4{,}093 \\
\bottomrule
\end{tabular}
\end{table}

\section{Annotation and Validation}

\subsection{Representation}

We use UD-style CoNLL-U fields for syntactic modeling: FORM, LEMMA, UPOS, FEATS, HEAD, and DEPREL. The repository schema (\repofile{configs/annotation_schema.yaml}) follows UD v2 categories and adds Katharevousa-oriented sidecar fields for archaic lexeme class, orthographic source, legacy morphology flags, and legal-register markers. These sidecar fields are not used directly in all model runs, but they document the linguistic phenomena that motivated the task: productive dative forms, infinitival residues, archaic prepositional government, participial constructions, formulaic openings, and institutional titles.

\subsection{LLM-assisted annotation protocol}

The annotation pipeline (\repofile{scripts/build_gpt_gold_dataset.py}, \repofile{scripts/resume_gpt_gold_batches.py}, \repofile{scripts/retry_failed_gold_dataset.py}) uses schema-constrained batch generation with offset-based progress tracking, resumable state, retry queues, conservative parsing of near-valid JSON, and validation before records are admitted to the frozen snapshot. This operational layer is central to the contribution. Historical low-resource annotation often fails through small unlogged decisions: one malformed batch, one changed split, or one silent retry can change downstream scores. Here, retry replacements are counted and the final manifest anchors the exact reference file used by all experiments.

The annotations are automatically validated rather than fully expert-adjudicated. The resource is therefore defined as a frozen reference annotation set: a stable benchmark snapshot suitable for controlled model comparison, error analysis, and targeted future human adjudication. Sampled philologist review is in progress and will be released as a versioned update of the same artifact.

\section{Models}

\subsection{External baselines}

We evaluate three off-the-shelf pipelines on the same held-out test split using \repofile{scripts/evaluate_external_baselines.py} and \repofile{scripts/evaluate_stanza_baseline.py}:

\begin{itemize}[leftmargin=1.5em]
  \item spaCy Greek \texttt{el\_core\_news\_lg};
  \item Stanza Greek pretrained pipeline \texttt{tokenize,pos,lemma,depparse};
  \item Stanza Ancient Greek PROIEL \texttt{grc\_PROIEL}.
\end{itemize}

Stanza is a neural multilingual NLP toolkit with pretrained pipelines for tokenization, tagging, lemmatization, and dependency parsing \cite{qi2020stanza}. All systems are evaluated against the same reference token boundaries and scored with the same evaluation code (\repofile{src/kathnlp/evaluation/metrics.py}). This isolates tagging and dependency decisions from sentence- and token-segmentation differences.

\subsection{Custom models}

The custom model set includes four families. A feature-based logistic parser (\repofile{src/kathnlp/training/models.py}, \repofile{scripts/train_and_evaluate.py}) uses enriched lexical-context features and learned head labels. An XLM-R parser (\repofile{src/kathnlp/training/transformer_parser.py}, \repofile{scripts/train_transformer_parser.py}) uses \texttt{xlm-roberta-base} with three training epochs, batch size 4, learning rate \(2 \times 10^{-5}\), weight decay 0.01, and maximum sequence length 256. An mBERT run uses \texttt{bert-base-multilingual-cased} with six epochs, batch size 8, warmup, and arc-emphasis loss weights. A custom Stanza model (\repofile{scripts/train_stanza_custom.py}) trains POS and dependency components with strict CoNLL-U handling and a 600-step training budget.

\begin{table}[H]
\centering
\caption{Primary model configurations.}
\label{tab:configs}
\begin{tabular}{p{2.8cm}p{3.0cm}p{7.0cm}}
\toprule
Family & Model & Configuration \\
\midrule
External & spaCy Greek & \texttt{el\_core\_news\_lg}, evaluated on fixed tokenized test split \\
External & Stanza Greek & \texttt{el} pretrained pipeline with POS, lemma, and dependency parsing \\
External & Stanza Ancient Greek & \texttt{grc\_PROIEL} pretrained pipeline with POS, lemma, and dependency parsing \\
Feature-based & Logistic baseline & Enriched lexical-context features and learned head labels \\
Transformer & XLM-R & \texttt{xlm-roberta-base}; 3 epochs; batch size 4; learning rate \(2 \times 10^{-5}\); max length 256 \\
Transformer & mBERT & \texttt{bert-base-multilingual-cased}; 6 epochs; batch size 8; warmup ratio 0.06; arc-emphasis loss weights \\
Stanza custom & \texttt{el\_kath} & Custom POS and depparse models; 600 steps; 10\% development split from training pool \\
\bottomrule
\end{tabular}
\end{table}

\section{Evaluation}

We report four metrics:

\begin{itemize}[leftmargin=1.5em]
  \item UPOS accuracy;
  \item weighted dependency-relation F1;
  \item UAS, the unlabeled attachment score;
  \item LAS, the labeled attachment score.
\end{itemize}

LAS is the primary structural measure because downstream syntax-sensitive analysis requires both the correct head and relation label. UPOS and DEPREL F1 are reported separately because they diagnose different error surfaces: the feature-based model, for example, is strong on local tagging and relation categories even when it does not produce the best attachment structure. All scoring is performed by the single evaluation module \repofile{src/kathnlp/evaluation/metrics.py}, with per-run JSON reports stored under \repodir{reports}.

\section{Results}

Table~\ref{tab:main} gives the fixed-split results. The best external baseline is spaCy Greek with 0.4183 LAS. Stanza Greek pretrained has a high UAS relative to its LAS, suggesting that it often attaches tokens to plausible heads while assigning labels that diverge from the reference annotation. Stanza Ancient Greek PROIEL performs worst overall, supporting the claim that Katharevousa parliamentary text is not simply an Ancient Greek transfer problem.

\begin{table}[H]
\centering
\caption{Main benchmark results on the fixed 340-sentence held-out split. Source reports: \repofile{reports/external_baselines_report.json}, \repofile{reports/stanza_baseline_report.json}, \repofile{reports/stanza_custom_v1_report.json}, \repofile{reports/transformer_parser_mbert_v2_report.json}, \repofile{reports/final_gold_eval_report.json}, \repofile{reports/transformer_parser_v1_opt_report.json}.}
\label{tab:main}
\begin{tabular}{lrrrr}
\toprule
Model & UPOS & DEPREL F1 & UAS & LAS \\
\midrule
spaCy Greek & 0.6721 & 0.5315 & 0.5492 & 0.4183 \\
Stanza Ancient Greek PROIEL & 0.5292 & 0.4044 & 0.4850 & 0.3076 \\
Stanza Greek pretrained & 0.6125 & 0.4242 & 0.6079 & 0.3396 \\
Stanza custom & 0.7694 & 0.6588 & 0.5756 & 0.4943 \\
mBERT & 0.8260 & 0.6076 & 0.5886 & 0.4537 \\
Logistic baseline & \textbf{0.9040} & \textbf{0.7451} & 0.5781 & 0.5072 \\
XLM-R & 0.8893 & 0.7250 & \textbf{0.6098} & \textbf{0.5162} \\
\bottomrule
\end{tabular}
\end{table}

\begin{figure}[H]
\centering
\includegraphics[width=\linewidth]{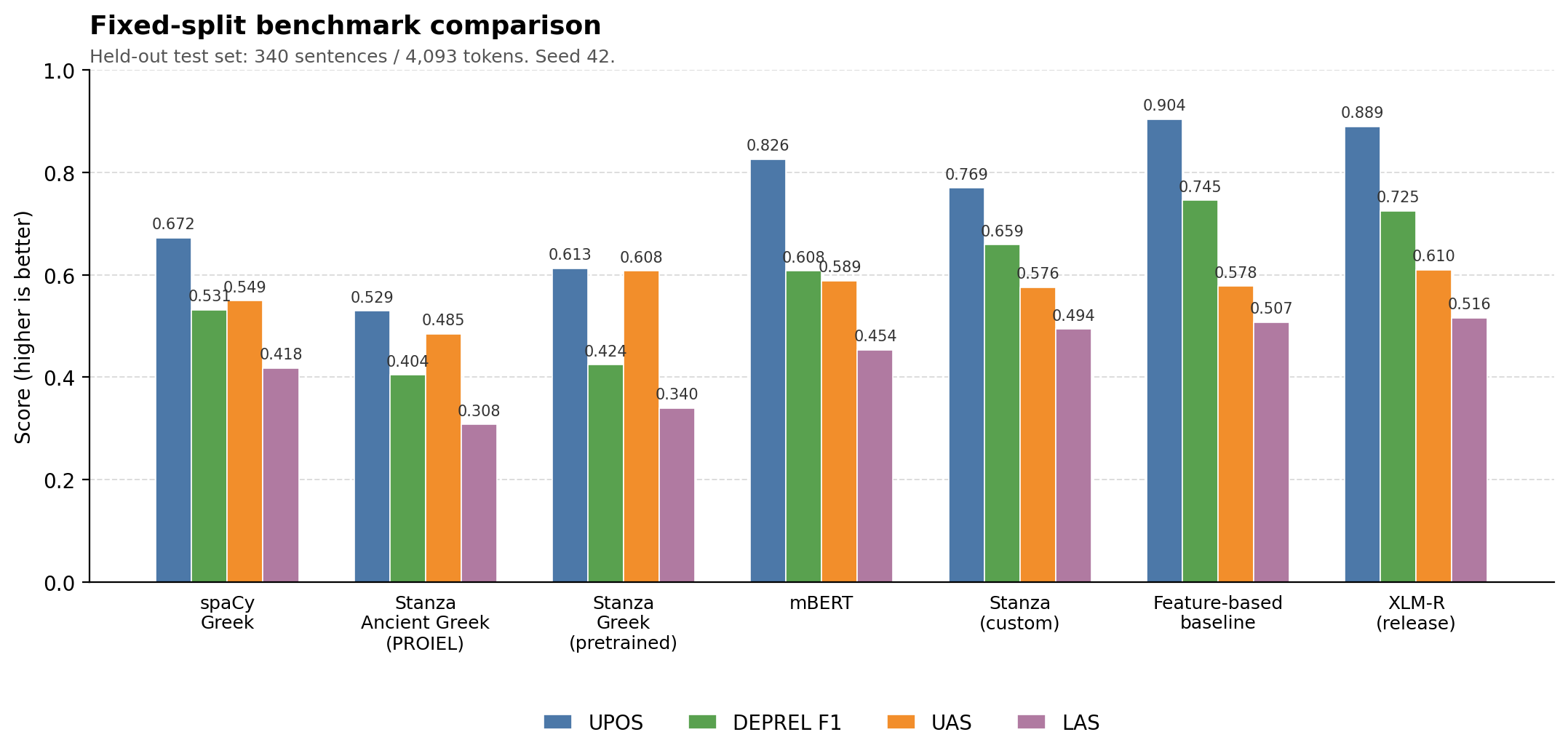}
\caption{Metric comparison across external baselines and custom models on the fixed held-out split.}
\label{fig:benchmark}
\end{figure}

The best structural parser is XLM-R, with 0.6098 UAS and 0.5162 LAS. Its LAS gain over the strongest external baseline is 0.0980 absolute, or approximately 23.4 percent relative to the baseline LAS. The feature-based logistic parser is a useful counterpoint: it has the best UPOS and DEPREL F1, and its LAS is only 0.0090 below XLM-R. This result argues against a simplistic ``transformers solve the task'' narrative. At the current data size, lexical-context features remain highly competitive for local morphosyntactic decisions, while XLM-R is stronger for attachment.

The custom Stanza model substantially improves over pretrained Stanza in LAS, from 0.3396 to 0.4943, even though it does not beat XLM-R. mBERT underperforms XLM-R on all four metrics, consistent with prior evidence that XLM-R often improves cross-lingual transfer under low-resource conditions \cite{conneau2020xlmr}. The margin is not enormous in UAS but is material in LAS, where relation-sensitive syntactic quality matters most.

\section{Open-Source Release and Reproducibility}

The pipeline is released as an open-source Python project at \url{\repobase}. The repository is organized around reproducible artifacts rather than a single monolithic script. The released workflow exports source files, reconstructs OCR-derived answer text, freezes the final reference snapshot, trains internal baselines, trains transformer parsers, and evaluates external baselines. Each empirical claim in this paper is anchored to a concrete file in the repository:

\begin{itemize}[leftmargin=1.5em]
  \item \repofile{data/processed/final_gold/gold_final.conllu}: frozen sentence-level reference annotations;
  \item \repofile{data/processed/final_gold/snapshot_manifest.json}: manifest for the frozen snapshot;
  \item \repofile{reports/transformer_parser_v1_opt_split.json}: fixed seed-42 train/test split;
  \item \repofile{reports/transformer_parser_v1_opt_report.json}: XLM-R release-candidate run;
  \item \repofile{reports/transformer_parser_mbert_v2_report.json}: mBERT comparison;
  \item \repofile{reports/stanza_custom_v1_report.json}: custom Stanza training report;
  \item \repofile{reports/stanza_baseline_report.json}: pretrained Stanza baselines;
  \item \repofile{reports/external_baselines_report.json}: spaCy Greek and Stanza external baselines;
  \item \repofile{reports/final_gold_eval_report.json}: feature-based logistic baseline.
\end{itemize}

The software environment is specified in \repofile{pyproject.toml}, which records the Python dependency envelope: pandas, OpenPyXL, python-docx, Pydantic, PyYAML, scikit-learn, NumPy, PyTorch, and Transformers. Random seeds and split files are stored alongside the benchmark reports. Hardware details for transformer training and model-weight availability are treated as release metadata in the public artifact package. A short maintainer guide for end-to-end corpus reconstruction is provided in \repofile{docs/MAINTAINER.md}.

To reproduce the XLM-R release candidate from a fresh clone:

\begin{quote}
\texttt{git clone \repobase} \\
\texttt{cd katharevousa-nlp-tooling} \\
\texttt{pip install -e .} \\
\texttt{python scripts/train\_transformer\_parser.py \textbackslash} \\
\texttt{\ \ --gold-path data/processed/final\_gold/gold\_final.conllu \textbackslash} \\
\texttt{\ \ --encoder-name xlm-roberta-base --epochs 3 --batch-size 4}
\end{quote}

\noindent External baselines can be reproduced under the same split with \repofile{scripts/evaluate_external_baselines.py}.

\section{Error Analysis and Optimization Path}

The released benchmark reports document several negative or mixed optimization tracks. Learning-rate schedule changes and loss reweighting produced broad regressions relative to the XLM-R \texttt{v1\_opt} configuration (\repofile{reports/transformer_parser_v2_opt_report.json}, \repofile{reports/transformer_parser_v2b_opt_report.json}). Hard-case augmentation was inconsistent (\repofile{reports/final_gold_eval_report_v4_hardcase.json}). Tokenizer truncation required strict old/new index mapping to avoid head-index errors. Stanza training exposed structural inconsistencies in portions of the data because the framework enforces UD tree constraints more strictly than the custom evaluation scripts.

These negative and mixed findings are part of the empirical result. In small historical treebanks, model selection can be distorted by accidental split drift, optimistic data cleaning, or unreported failed runs. The negative results show that the selected release candidate is not the first model tried, but the best configuration within a documented search space. They also identify where future work is likely to pay off: longer-distance attachments, relation labeling under archaic prepositional government, and explicit modeling of sentence length and enumerative constructions.

\section{Discussion}

The experiments support three conclusions. First, Katharevousa parliamentary parsing is a real register-transfer problem. Modern Greek systems have some useful lexical and structural coverage, but their labeled attachment performance remains below custom models. Ancient Greek transfer is weaker, presumably because the parliamentary text combines archaizing grammar with modern institutional vocabulary and postwar administrative formulae.

Second, the strongest model family depends on the metric. XLM-R is the best release candidate for structural parsing, but the logistic baseline is strongest for UPOS and dependency-relation classification. This pattern points toward hybrid modeling strategies that combine transformer representations with explicit register-aware features rather than treating the two as mutually exclusive.

Third, reproducibility is itself a methodological contribution. The released repository records a complete path from OCR reconstruction to frozen CoNLL-U snapshot, fixed split, model reports, and external baselines. This matters for both NLP and digital humanities: the same infrastructure that supports parser benchmarking also supports transparent reuse of parliamentary archives for historical, political, and linguistic questions.

\section{Limitations}

The current reference set is small: 1{,}697 sentences, with 340 sentences in the held-out split. The results are release-candidate benchmarks rather than final performance ceilings for Katharevousa parsing. The data come from parliamentary questions in 1976--1977, so transfer to other Katharevousa genres, earlier legal texts, newspapers, or mixed polytonic sources remains untested. The labels have been automatically validated, not fully expert-adjudicated; this limits claims about absolute linguistic correctness and makes sampled human audit a necessary next validation step. Finally, all external systems were evaluated with reference token boundaries; end-to-end performance from raw OCR text will be lower and requires separate measurement.

\section{Ethics and Data Statement}

The source documents are historical parliamentary records. They concern public political activity rather than private communication, but provenance, licensing, and archival context remain important for reuse. LLM assistance was used as constrained infrastructure, not as an autonomous historical authority. The release package distinguishes generated annotations, automatically validated outputs, manually checked components, and artifacts intended for downstream scholarly use.

\section{Conclusion}

We presented a reproducible dependency parsing pipeline for Katharevousa Greek parliamentary text and released it as an open-access companion library, \texttt{kathnlp}. The workflow reconstructs OCR-derived source material, produces a frozen UD-style reference snapshot, and evaluates external and custom parsers under a fixed split. The best structural parser is an XLM-R model with 0.5162 LAS, outperforming the strongest off-the-shelf baseline by 0.0980 absolute LAS. The feature-based parser remains highly competitive for UPOS and relation labeling, showing that transparent linguistic features still matter in this low-resource historical setting. The main contribution is therefore both empirical and infrastructural: a first release candidate for Katharevousa syntactic parsing, and an auditable workflow for converting difficult archival text into reusable NLP resources. We invite the historical NLP and digital-humanities communities to extend the resource through expert adjudication, register expansion, and downstream applications, building on the released code and benchmark protocol at \url{\repobase}.

\section*{Code and Data Availability}

All artifacts associated with this paper are openly available in the project repository at \url{\repobase}, including:

\begin{itemize}[leftmargin=1.5em]
  \item the Python package \texttt{kathnlp} under \repodir{src/kathnlp};
  \item the data construction, training, and evaluation scripts under \repodir{scripts};
  \item the annotation schema in \repofile{configs/annotation_schema.yaml};
  \item the frozen reference treebank, manifest, and fixed split under \repodir{data/processed/final_gold} and \repodir{reports};
  \item the LaTeX source of this manuscript under \repodir{paper}.
\end{itemize}

\noindent License terms for code, annotations, and benchmark reports are stated in the repository \href{\repobase/blob/main/README.md}{\texttt{README.md}}. The XLM-R release-candidate model weights are planned for distribution on the Hugging Face Hub under the project name \texttt{kathnlp}. Issues, contributions, and adjudication proposals are welcome via the repository's issue tracker.

\bibliographystyle{plain}
\bibliography{references}

\begin{thebibliography}{10}

\bibitem{conneau2020xlmr}
Alexis Conneau, Kartikay Khandelwal, Naman Goyal, Vishrav Chaudhary, Guillaume Wenzek, Francisco Guzm{\'a}n, Edouard Grave, Myle Ott, Luke Zettlemoyer, and Veselin Stoyanov.
\newblock Unsupervised cross-lingual representation learning at scale.
\newblock In {\em Proceedings of the 58th Annual Meeting of the Association for Computational Linguistics}, pages 8440--8451. Association for Computational Linguistics, 2020.

\bibitem{devlin2019bert}
Jacob Devlin, Ming-Wei Chang, Kenton Lee, and Kristina Toutanova.
\newblock {BERT}: Pre-training of deep bidirectional transformers for language understanding.
\newblock In {\em Proceedings of the 2019 Conference of the North American Chapter of the Association for Computational Linguistics: Human Language Technologies}, pages 4171--4186. Association for Computational Linguistics, 2019.

\bibitem{duong2015lowresource}
Long Duong, Trevor Cohn, Steven Bird, and Paul Cook.
\newblock A neural network model for low-resource universal dependency parsing.
\newblock In {\em Proceedings of the 2015 Conference on Empirical Methods in Natural Language Processing}, pages 339--348. Association for Computational Linguistics, 2015.

\bibitem{fitsilis2024digitization}
Fotios Fitsilis, Basilis Gatos, Konstantinos Palaiologos, Panagiotis Kaddas, Charalambis Kyrkos, Maria-Eleni Georgoulea, Yiannis Armenakis, Christina Tasouli, George Mikros, Olivier Rozenberg, and Eleni Kiousi.
\newblock Digitization of written parliamentary questions from the historical archive (1974--1977) of the hellenic parliament.
\newblock In H.~Mouch{\`e}re and A.~Zhu, editors, {\em Document Analysis and Recognition -- ICDAR 2024 Workshops}, volume 14935 of {\em Lecture Notes in Computer Science}, pages 103--117, Cham, 2024. Springer Nature Switzerland.

\bibitem{kondratyuk2019udify}
Dan Kondratyuk and Milan Straka.
\newblock 75 languages, 1 model: Parsing universal dependencies universally.
\newblock In {\em Proceedings of the 2019 Conference on Empirical Methods in Natural Language Processing and the 9th International Joint Conference on Natural Language Processing}, pages 2779--2795. Association for Computational Linguistics, 2019.

\bibitem{mackridge2009}
Peter Mackridge.
\newblock {\em Language and National Identity in Greece, 1766--1976}.
\newblock Oxford University Press, 2009.

\bibitem{makarov2020normalization}
Peter Makarov and Simon Clematide.
\newblock Semi-supervised contextual historical text normalization.
\newblock In {\em Proceedings of the 58th Annual Meeting of the Association for Computational Linguistics}, pages 7284--7295. Association for Computational Linguistics, 2020.

\bibitem{mikrosfitsilis2025dsh}
George Mikros and Fotios Fitsilis.
\newblock Computational stylistics of post-junta greek parliamentary questions: Katharevousa, party style, and democratic reconstruction (1976--1977), 2025.
\newblock Manuscript submitted to Digital Scholarship in the Humanities.

\bibitem{nivre2020ud}
Joakim Nivre, Marie-Catherine de~Marneffe, Filip Ginter, Jan Haji{\v{c}}, Christopher~D. Manning, Sampo Pyysalo, Sebastian Schuster, Francis Tyers, and Daniel Zeman.
\newblock Universal dependencies v2: An evergrowing multilingual treebank collection.
\newblock In {\em Proceedings of the Twelfth Language Resources and Evaluation Conference}, pages 4034--4043. European Language Resources Association, 2020.

\bibitem{qi2020stanza}
Peng Qi, Yuhao Zhang, Yuhui Zhang, Jason Bolton, and Christopher~D. Manning.
\newblock Stanza: A python natural language processing toolkit for many human languages.
\newblock In {\em Proceedings of the 58th Annual Meeting of the Association for Computational Linguistics: System Demonstrations}, pages 101--108. Association for Computational Linguistics, 2020.

\bibitem{richter2018ocr}
Caitlin Richter, Matthew Wickes, Deniz Beser, and Mitch Marcus.
\newblock Low-resource post processing of noisy {OCR} output for historical corpus digitisation.
\newblock In {\em Proceedings of the Eleventh International Conference on Language Resources and Evaluation}. European Language Resources Association, 2018.

\bibitem{udformat}
{Universal Dependencies}.
\newblock {CoNLL-U} format.
\newblock \url{https://universaldependencies.org/format.html}, 2026.
\newblock Accessed 2026-05-05.

\bibitem{wu2018multilingual}
Yingting Wu, Hai Zhao, and Jia-Jun Tong.
\newblock Multilingual universal dependency parsing from raw text with low-resource language enhancement.
\newblock In {\em Proceedings of the {CoNLL} 2018 Shared Task: Multilingual Parsing from Raw Text to Universal Dependencies}, pages 74--80. Association for Computational Linguistics, 2018.

\end{thebibliography}

\end{document}